\documentclass[10pt,twocolumn,letterpaper]{article}

\usepackage{iccv}              

\definecolor{iccvblue}{rgb}{0.21,0.49,0.74}
\usepackage[pagebackref,breaklinks,colorlinks,allcolors=iccvblue]{hyperref}

\usepackage{times}
\usepackage{latexsym}

\usepackage{amsmath}
\usepackage{graphicx}
\usepackage{float}
\usepackage{amsmath,amssymb}

\usepackage{amsmath}
\usepackage{float}
\usepackage{algorithm}
\usepackage{algorithmic}
\usepackage{amsmath,amssymb}
\usepackage{dsfont}
\usepackage{multirow}
\usepackage{fancyhdr}

\usepackage[T1]{fontenc}

\def\paperID{*****} 
\def\confName{ICCV}
\def\confYear{2025}

\title{Efficient Learning for Product Attributes with Compact Multimodal Models}

\author{Mandar Kulkarni\\
Flipkart Data Science\\
Seattle, USA\\
{\tt\small mandar.kulkarni@flipkart.com}
}

\begin{document}

\maketitle

\begin{abstract}

Image-based product attribute prediction in e-commerce is a crucial task with numerous applications. The supervised fine-tuning of Vision Language Models (VLMs) faces significant scale challenges due to the cost of manual or API based annotation. In this paper, we investigate label-efficient semi-supervised fine-tuning strategies for compact VLMs (2B–3B parameters) that leverage unlabeled product listings through Direct Preference Optimization (DPO). Beginning with a small, API-based, annotated, and labeled set, we first employ PEFT to train low-rank adapter modules. To update the adapter weights with unlabeled data, we generate multiple reasoning-and-answer chains per unlabeled sample and segregate these chains into preferred and dispreferred based on self-consistency. We then fine-tune the model with DPO loss and use the updated model for the next iteration. By using PEFT fine-tuning with DPO, our method achieves efficient convergence with minimal compute overhead. On a dataset spanning twelve e-commerce verticals, DPO-based fine-tuning, which utilizes only unlabeled data, demonstrates a significant improvement over the supervised model. Moreover, experiments demonstrate that accuracy with DPO training improves with more unlabeled data, indicating that a large pool of unlabeled samples can be effectively leveraged to improve performance.



\end{abstract}

\thispagestyle{fancy}
\fancyhf{}
\rhead{Accepted at International Conference on Computer Vision (ICCV) 2025 Workshop on Curated Data for Efficient Learning 
}

\section{Introduction}

\begin{table*}[t]
\centering
\begin{tabular}{|p{1.5cm}|p{2.5cm}|p{5cm}|p{6cm}|}
\hline
\textbf{Vertical} & \textbf{Attribute} & \textbf{Description} & \textbf{Allowed values} \\
\hline
suitcase & pattern & Pattern refers to a repeated decorative sketch that is printed or engraved on a product. & 

Character,Checkered,Colorblock,Printed,  Solid,   Striped,Textured,Abstract,Tropical,  Geometric-Print,Graphic-Print,Floral-Print,Polka-Print, Animal-Print\\
\hline
kurta &  detail placement & Detail Placement refers to the print or design placed on particular place of the garment. & All-Over, Back, Waist, Hemline, Neckline, None,Side, Sleeve, Slits, Yoke,Chest,Cuff,Pocket,Front-Panel\\
\hline

\end{tabular}
\caption{Examples of vertical specific attribute, description and allowed values}
\label{tab:vad}

\end{table*}

\begin{figure*} [!t]
\centering
\begin{tabular}{c c c c}

\includegraphics[width=110pt, height = 160pt]{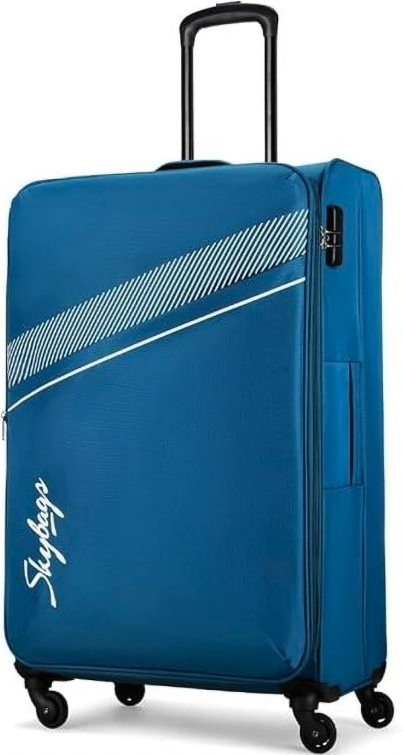}&
\includegraphics[width=120pt, height = 130pt]{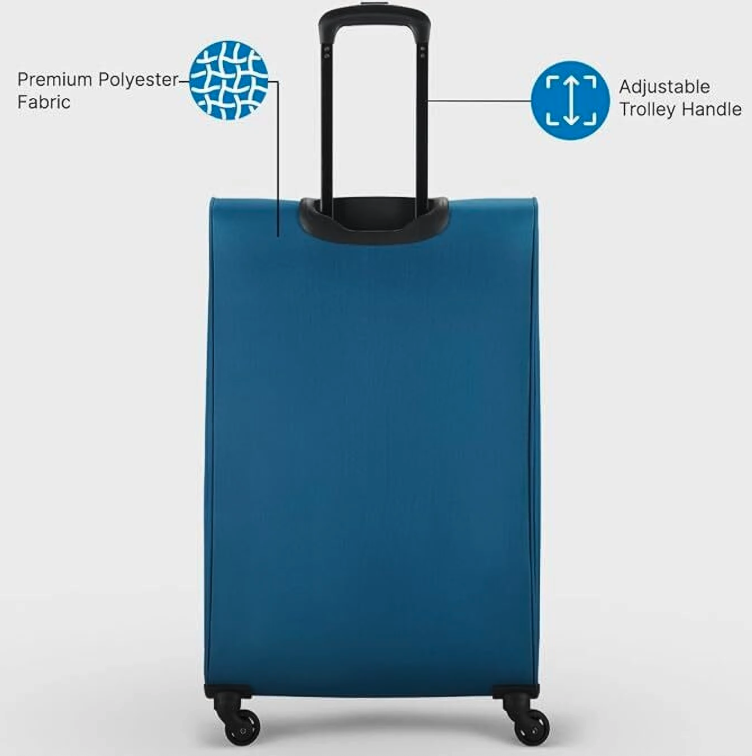}&
\includegraphics[width=110pt, height = 140pt]{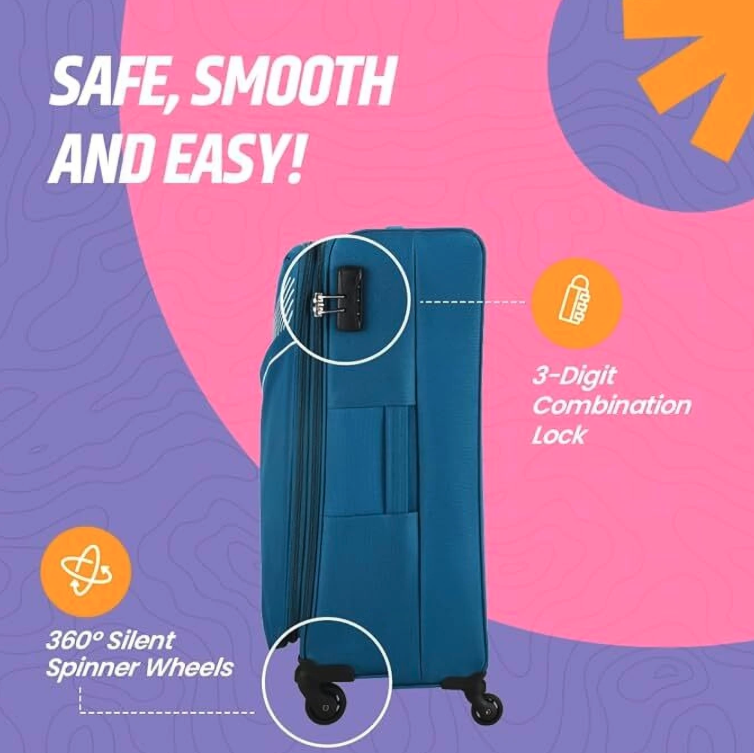}\\

\end{tabular}
\caption{Images from the suitcase vertical. Typically e-commerce images contain various views of the same product at different scales.}
\label{fig:suticase}
\end{figure*}

Image-based attribute prediction is a crucial task in the e-commerce domain, with numerous practical applications, including automated listing assistance and quality control. Unlike open‑ended generation tasks, attribute prediction is inherently objective: the values produced by the model can be directly verified against ground truth or image content.

Vision–Language Models (VLMs) have recently achieved state-of-the-art performance on cross-modal benchmarks by fusing visual features with textual embeddings. However, large-scale multimodal architectures incur prohibitive computational and memory costs, which restrict their use in latency-sensitive applications common in e-commerce operations.

Supervising the fine-tuning of such large VLMs is impractical in the e-commerce setting. Modern catalogs host millions of items across diverse categories—or “verticals”—and manual labeling at this scale is infeasible. Moreover, seller‑supplied attribute values (e.g., “color,” “occasion,” “ideal\_for”) are often noisy and require extensive manual quality control. Although large VLM APIs can generate labels automatically, the operational expense becomes prohibitive at catalog scale. In contrast, unlabeled images from product listings are abundant and represent a valuable resource if harnessed effectively.

In this work, we investigate the viability of small VLMs (2–3 billion parameters) for image‑based attribute prediction in e‑commerce. We evaluate two Qwen-family models—Qwen2‑VL‑2B‑Instruct and Qwen2.5‑VL‑3B‑Instruct—by adapting them to the attribute-prediction task.
To minimize computational overhead, we employ Parameter-Efficient Fine-Tuning (PEFT) \cite{peft}, updating only the low-rank adapter weights. We first use large VLM APIs to generate reasoning‑and‑attribute labels for a small subset of product listings. We fine‑tune adapters on this labeled seed set, then perform adapter tuning on 5,000 unlabeled listings (each with up to five images) using the DPO objective. 

Prior research has shown that self-consistency strongly correlates with model uncertainty and accuracy on chain-of-thought tasks \cite{wang2023selfconsistencyimproveschainthought, Trad_2025}. Building on this insight, we explore the Direct Preference Optimization (DPO) approach for Unsupervised Fine-Tuning (UFT) using unlabeled data. DPO \cite{rafailov2023direct} replaces conventional reinforcement learning objectives with a convex loss that directly aligns model log probabilities with observed preference data. For each unlabeled example, we generate multiple reasoning-and-answer chains and identify the most consistent prediction to designate "preferred" versus "dispreferred" outputs. We then fine-tune the model using DPO loss on these pseudo-labels. We iterate this process by regenerating pseudo-labels with the updated model, followed by DPO fine-tuning. To avoid updates based on predictions of high uncertainty, we employ a consistency confidence threshold, defined as the fraction of predictions that match the most consistent prediction. The model is updated only for predictions that exceed this threshold.

For training with unlabeled data, we also experimented with the Self-Learning (SL) objective, which showed promising results on high-capacity LLMs \cite{huang2022largelanguagemodelsselfimprove}.
We implement a pseudo-labeling loop in which the VLM generates candidate labels for unlabeled listings, filters them based on the confidence threshold, and retrains itself only on these preferred generations. However, we observed that the SL objective degrades the accuracy in small VLM settings.

Our train and test datasets span twelve e-commerce verticals, each defined by a set of visual attributes (e.g., color, pattern, occasion, ideal\_for). Experimental results demonstrate that DPO‑based fine‑tuning on pseudo‑labeled data yields significant improvements in attribute‑prediction accuracy. 
The high accuracy lifts are observed even when pretrained checkpoints are directly updated using DPO loss on unlabeled data.
We observe that the accuracy with DPO training improves with an increasing amount of unlabeled data, indicating that a large pool of unlabeled samples can be effectively leveraged to enhance performance.

\section{Data description}

\begin{table*}[!t]
    \centering
    \captionsetup{justification=centering, margin=5mm}
    \begin{tabular}{|p{3cm}|p{5.5cm}|p{6.5cm}|}
        
        \hline
        \textbf{Attribute} & \textbf{Question prompt} & \textbf{Answer prompt} \\
        \hline
        
        type & what is the most suitable value for \textbf{type} for this \textbf{suitcase}.
Type refers to the classification of the product based on its use and structure.Options: ['Cabin \& Check-in Set', 'Check-in Suitcase', 'Cabin Suitcase', 'Overnighter \& Briefcase'] & <reasoning>The suitcase appears to be a standard check-in suitcase based on its size and design, which is suitable for larger travel needs.</reasoning><answer>Check-in Suitcase</answer> \\  
        \hline
        locking\_mechanism & what is the most suitable value for \textbf{locking\_mechanism} for this \textbf{suitcase}.
Locking Mechanism refers to the locking mechanism of the product.Options: ['None', 'Number Lock', 'TSA Lock', 'Key Lock', 'Combination Lock', 'Smart Lock']
 & <reasoning>The suitcase features a 3-digit combination lock as indicated in the third image, which clearly shows the locking mechanism.</reasoning><answer>Combination Lock</answer> \\

        \hline
        
        \end{tabular}
    \vspace{2mm}
\caption{Sample training examples, prepared using prompt formats, for two attributes of a listing from Suitcase vertical}
    \label{tab:te}
\end{table*}

\begin{table}[t]
    \centering
    \captionsetup{justification=centering, margin=5mm}
    \begin{tabular}{p{7cm}}
                       
        \hline
        
        \textbf{System Prompt}:
        
        Respond in the following format:<reasoning>  ...  </reasoning><answer>  ...     </answer>\\
    
        \vspace{0.2cm}
        
        \textbf{User Prompt}:
        
        what is the most suitable value for <attribute name> for >this <vertical>. <attribute description> Options: <allowed values> \\

        \vspace{0.2cm}

        \textbf{Answer prompt}:
        
        <reasoning>  reason  </reasoning><answer>  answer  </answer> 
        
        \\ \hline

        \end{tabular}
    \vspace{2mm}
    \caption{Prompt formats used for the VLM training}
    \label{tab:pf}
\end{table}

E‑commerce catalog data typically comprises multiple product images captured from different viewpoints, organized hierarchically by product category (or “vertical”) down to individual listings. Each vertical defines its own set of attributes that are displayed on the product page—for example, footwear verticals like “sandal” or “slipper flip‑flop” may include attributes such as heel pattern or color, whereas the “suitcase” vertical has attributes such as locking mechanism, number of wheels. For our experiments, we selected product listings from twelve distinct verticals—sandal, ethnic set, kurta, shoe, slipper flip‑flop, trouser, t‑shirt, top, shirt, suitcase, sari, and watch. To help sellers list products efficiently, e-commerce platforms provide vertical-dependent attribute descriptions and sets of allowed values, enabling quick and standardized input. In curating our dataset, we utilized seller‑provided images alongside each attribute’s description and allowed values. Figure \ref{fig:suticase} illustrates example product images for a listing in the suitcase vertical, while Table \ref{tab:vad} presents examples of vertical specific attributes, descriptions, and allowed value sets.

For unlabeled data, we selected 5,000 listings (each with up to five images) across these twelve verticals. To bootstrap supervised training, we curated a small labeled dataset by randomly sampling ten listings per vertical (627 total attribute instances) and annotating them via a GPT–4o–mini–based pipeline. In this pipeline, the GPT‑4o‑mini multimodal model receives product images along with the attribute name, description, and allowed values and is prompted to generate both the reasoning chain and the final attribute label. 

For rigorous evaluation, we assembled a held-out test set consisting of 3,175 listings with 7,145 attribute annotations, each of which was manually verified for correctness. This carefully curated test set enables us to measure how well a small VLM generalizes across the diverse visual conditions found in real-world e-commerce imagery, including varied backgrounds, inconsistent lighting, and the presence of people or ancillary objects. By benchmarking against this dataset, we can assess both the overall attribute-prediction accuracy and the model’s ability to localize to the correct image regions, ultimately demonstrating the feasibility of utilizing compact, efficient VLMs for large-scale product attribute prediction.

\section{Proposed methodology}

In this section, we provide details of the VLM training approach. For all the experiments, input images are resized to 224 x 224 pixels. For test data inference, greedy decoding is used with a maximum of 200 decoding tokens.




\begin{table*}[!t]
\centering
\begin{tabular}{|l|l|l|l|l|}
\hline
\textbf{Technique} & \textbf{Setting} & \textbf{Model} & \textbf{PT} & \textbf{Peft-10} \\
\hline
PEFT & SFT & Qwen2-VL-2B-Instruct & 0.317 & 0.44  \\
\hline
PEFT & SFT & Qwen2.5-VL-3B-Instruct & 0.717 & 0.751 \\
\hline
SL & UFT & Qwen2.5-VL-3B-Instruct & 0.642 & 0.665 \\
\hline
DPO & UFT & Qwen2.5-VL-3B-Instruct & \textbf{0.826} & \textbf{0.857} \\
\hline

\end{tabular}
\caption{Results with Supervised Fine-Tuning (SFT) and Unsupervised Fine-Tuning (UFT). For SFT, PT indicates accuracy results on the test set with the pretrained checkpoint (i.e., without fine-tuning) while Peft-10 indicates results for the adapter trained from 10 listings per vertical. For UFT with Direct Preference Optimization (DPO) and Self Learning (SL), PT and Peft-10 models are used as the initial SFT checkpoints.}
\label{tab:con}

\end{table*}

\subsection{Supervised fine-tuning}

To obtain improved initializations for training on unlabeled data, we first fine-tune a small VLM on a labeled dataset obtained using the GPT-4o-mini pipeline, using Parameter Efficient Fine-Tuning (PEFT) \cite{peft}. The labeling can introduce annotation noise; however, PEFT has been shown to maintain robustness under such noisy supervision \cite{kim2024clearrobustgeneralizedparameterefficient}.  
The labeled dataset consists of 10 listings per vertical (a total of 120 listings with 627 attributes), with a generated reasoning and answer per attribute.

Our fine-tuning protocol utilizes a fixed system prompt that directs the VLM to generate a structured chain of reasoning, followed by its final attribute prediction. This is complemented by a user prompt specifying the product vertical, the attribute name, its textual description, and the set of allowed values. The model’s next‑token target incorporates both the reasoning trace and the final attribute label, formatted within a standardized “answer prompt” template. Table \ref {tab:pf} details the system, user, and answer prompt schemas, while Table \ref {tab:te} provides exemplar training instances created using this approach. 

For adapter tuning, we configure LoRA modules with rank 32 and alpha 32, updating only the query and value projection adapters while freezing the remaining model parameters. We conduct experiments on two Qwen‑family backbones—Qwen2‑VL‑2B‑Instruct and Qwen2.5‑VL‑3B‑Instruct, resulting in approximately 2.5M and 7.3M trainable parameters, respectively. Using a fixed learning rate of 3e-5, we fine‑tune each model on a labeled dataset.




Table \ref{tab:con} presents the PEFT supervised fine-tuning accuracy results on the test set using direct pre-trained checkpoints (without PEFT training) and their PEFT-fine-tuned version (trained on 10 listings per vertical). The pretrained Qwen2.5‑VL‑3B‑Instruct outperforms the Qwen2‑VL‑2B‑Instruct, and supplementing with a small labeled set yields further gains. We use Qwen2.5-VL-3B-Instruct pretrained and peft‑adapted models as Supervised Finetuned (SFT) baselines for our subsequent experiments, leveraging unlabeled data.

\subsubsection{Effect of reason generation}

Although attribute prediction is not inherently a logical reasoning task, we hypothesized that generating an explicit reasoning trace might still enhance model performance. To verify this, we conducted an ablation in which the VLM was trained solely on the final attribute answer—omitting the reasoning chain—by changing the answer prompt to <answer>{answer}</answer>. Inference likewise generated only the answer token. Under this configuration, Qwen2.5-VL-3B-Instruct, fine-tuned on 10 listings per vertical, achieved an accuracy of 0.607, significantly lower than when trained with the reasoning prompt. This result underscores the utility of reasoning generation, even for straightforward “show-and-tell” attribute tasks.

\section{Training with unlabeled data}
To leverage unlabeled product listings, we experiment with Direct Preference Optimization (DPO) initialized from the SFT baseline (pretrained checkpoint and PEFT‑adapted variant). 
For training with the PEFT variant, we further refine the existing adapter weights using DPO. In contrast, for training directly with the pre-trained checkpoint, we introduce new LoRA adapter modules and update them with DPO on the unlabeled data. 
In both cases, we freeze the core VLM parameters and update only the adapter weights.

Using 5000 unlabeled listings, each including up to five product images, we prompt the SFT model to generate seven complete reasoning‑and‑answer chains. The temperature value is set to 1 for generation. From these outputs, we isolate the answer tokens and apply self-consistency as a proxy for ground truth: the answer choice that appears most frequently across the generated chains becomes the “consensus” label. We then tag each generation as “Preferred” if it matches the consensus and “Dispreferred” otherwise. A consistency confidence score ($C$) is computed per attribute as the fraction of predictions that match the consensus label. The model is trained only on samples that exceed a 50\% confidence threshold, thereby filtering out highly uncertain pseudo-labels. We investigated the impact of $C$ on accuracy and found that setting this threshold to 50\% optimally balances label quality and data volume, resulting in the best downstream accuracy in our experiments.


\subsection{Details of DPO training}

To fine‑tune the small VLM using DPO, we construct pairwise training instances by randomly pairing each dispreferred generation with one preferred generation drawn from the same example’s preferred set. In this setup, the supervised fine‑tuned (SFT) version of the model serves as the fixed reference policy, while the current model parameters define the trainable policy. During each training step, we train with the DPO loss, which directly aligns the difference in sequence-level log probabilities between the preferred and dispreferred outputs, thereby favoring the former. Concretely, for a given prompt x, preferred completion $y^+$, and dispreferred completion $y^-$, we compute the DPO loss ($\mathcal{L}$) under the policy $\pi_{\theta}$ and under the reference $\pi_{\mathrm{ref}}$.

\begin{equation*}
\begin{split}
\mathcal{L}
&= -\mathbb{E}\Bigl[\log \sigma\Bigl(\beta\Bigl[\log \frac{\pi_{\theta}(y^+|x)}{\pi_{\mathrm{ref}}(y^+|x)} \;-\; \log \frac{\pi_{\theta}(y^-|x)}{\pi_{\mathrm{ref}}(y^-|x)}\Bigr]\Bigr)\Bigr] \\
&\quad 
\end{split}
\end{equation*}

We set the $\beta$ parameter to 0.2 and the learning rate to 1e-5. 
Table \ref{tab:con} presents the DPO accuracy results on the test set. Training with unlabeled data using DPO yields substantial gains in attribute prediction accuracy. For instance, the Peft-10 model improves from 0.751 under SFT alone to 0.857 after DPO training, yielding \textasciitilde{10\%} increase. Note that even directly fine-tuning a pre-trained checkpoint with unlabeled data provides \textasciitilde{11\%} improvement in the accuracy.
This indicates that any SFT VLMs can achieve significant performance enhancements through unlabeled preference-based fine-tuning. Table \ref{tab:slbdpo} in the shows examples of the predictions where DPO-based training on unlabeled samples has improved the accuracy of the SFT baseline. 





\subsubsection{Effect of sample confidence}

To quantify the effect of consistency confidence score ($C$) on DPO fine‑tuning, we conducted a controlled study in which the consistency threshold for selecting generation pairs was systematically varied. Starting from our Peft-10 SFT baseline, we generated seven complete reasoning‑and‑answer chains for each unlabeled example and computed the consensus confidence as the proportion of chains matching the most frequent answer. We then swept the consistency threshold ($C$) from 0.8 down to 0.3, retaining only those examples whose consensus confidence exceeded $C$ to form Preferred–Dispreferred pairs used in the DPO objective. Table \ref{tab:dpoc} shows the result. As anticipated, lowering the threshold ($C$) enlarged the training set but also admitted examples with higher uncertainty and potential label errors. At ($C = 0.8$), DPO has access to only a small yet high‑precision subset, thereby not achieving the highest lift. Reducing ($C$) to 0.7 increased data volume and yielded further accuracy gains, indicating that DPO’s convex preference loss can tolerate moderate pseudo‑label noise. Conversely, setting ($C = 0.3$) starts introducing excessively low‑confidence examples, resulting in performance degradation. In the attribute-prediction context, a threshold of ($C = 0.5$) worked the best. This underscores the critical role of confidence cutoff selection: higher values of ($C$) preserve label fidelity at the expense of data volume. In comparison, lower values expand training examples but risk noise‑induced errors.

\begin{table}[h]
\centering
\begin{tabular}{|l|l|l|l|}
\hline
\textbf{C=0.3} & \textbf{C=0.5} & \textbf{C=0.7} & \textbf{C=0.8}\\
\hline
0.83 & \textbf{0.857} & 0.831 & 0.825\\
\hline

\end{tabular}
\caption{Effect of confidence ($C$) on the DPO training}
\label{tab:dpoc}
\end{table}






\section{Self-Learning (SL)}

For training with unlabeled data, we also experimented with self-learning (SL). Huang et al. \cite{huang2022largelanguagemodelsselfimprove} have demonstrated promising results in improving accuracy with high-capacity language models by training with pseudo labels. 
Starting with the Peft-10 SFT model, we first generate seven reasoning‑and‑answer chains using a temperature value of 1. Using the samples above the confidence threshold of 0.5, we use a consensus label to identify preferred generations and use them as hard pseudo-labels in a next-token prediction task. After this fine-tuning step, we regenerate pseudo-labels using the updated model and repeat the process. Table \ref{tab:con} presents the SL accuracy results. We observe that self-learning degrades accuracy. This could be because a small VLM, with its limited capacity, generates highly similar reasoning and answer chains. Fine-tuning on these low-variance pseudo-labels concentrates the model’s probability mass on the same few tokens (model collapse), reinforcing any early mistakes. As a result, the consensus labels become increasingly reflective of the model’s biases rather than true answers, amplifying errors over iterations and degrading performance.

\section{Effect of number of unlabeled samples}

We conducted a controlled study to evaluate the impact of the volume of unlabeled product listings on model performance under both DPO and SL paradigms. Starting from our Peft-10 SFT baseline and setting $C$ to 0.5, we varied the size of the unlabeled pool from 1000 to 5000 listings and measured the resulting attribute-prediction accuracy for each method. As shown in Table \ref{tab:sldpo}, accuracy under the SL initially improves with more pseudo-labeled data but then degrades as additional unlabeled samples introduce greater label noise and reinforce model biases. In contrast, DPO training yields steadily increasing accuracy across the entire range. These findings confirm that, while simple self-learning can benefit from moderate amounts of unlabeled data, the DPO framework more effectively harnesses large-scale unlabeled corpora to drive continuous gains in prediction accuracy.

\begin{table}[h]
\centering
\begin{tabular}{|l|l|l|l|l|l|}
\hline
\textbf{setting} & \textbf{1k} & \textbf{2k} & \textbf{3k} & \textbf{4k} & \textbf{5k}\\
\hline
SL & 0.785 & 0.755 & 0.73 & 0.731 & 0.66 \\
\hline
DPO & 0.816 & 0.836 & 0.84 & 0.852 & 0.857\\
\hline
\end{tabular}
\caption{Effect of the number of unlabeled samples on SL and DPO fine-tuning. Note that accuracy with DPO improves with an increasing number of unlabeled samples. }
\label{tab:sldpo}
\end{table}

\begin{table*}[!t]
\centering
\begin{tabular}{|p{5cm}|p{5cm}|p{5cm}|}
\hline
\textbf{Product images} & \textbf{(Incorrect) PEFT prediction} & \textbf{(Corrected) DPO prediction}\\
\hline
\includegraphics[width=100pt, height = 120pt]{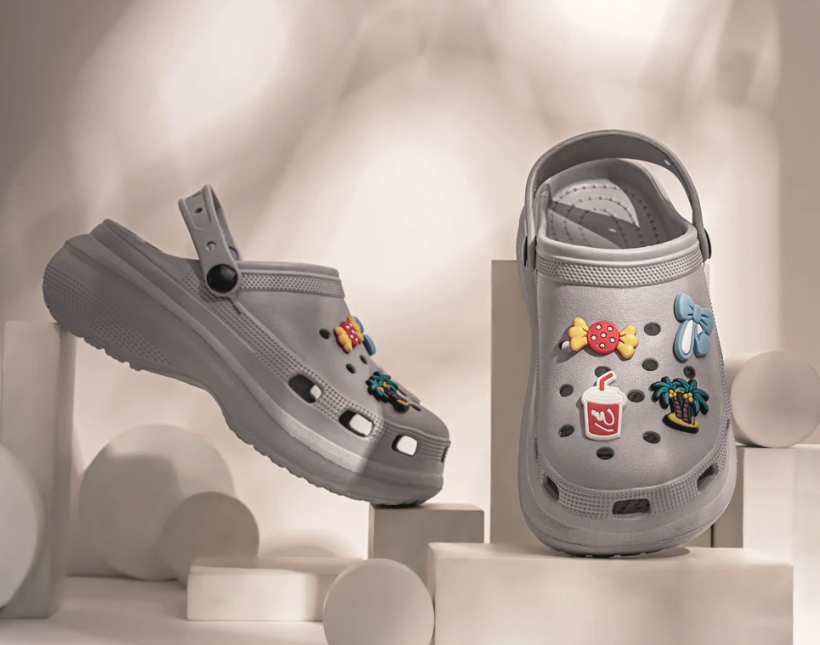}  \includegraphics[width=100pt, height = 120pt]{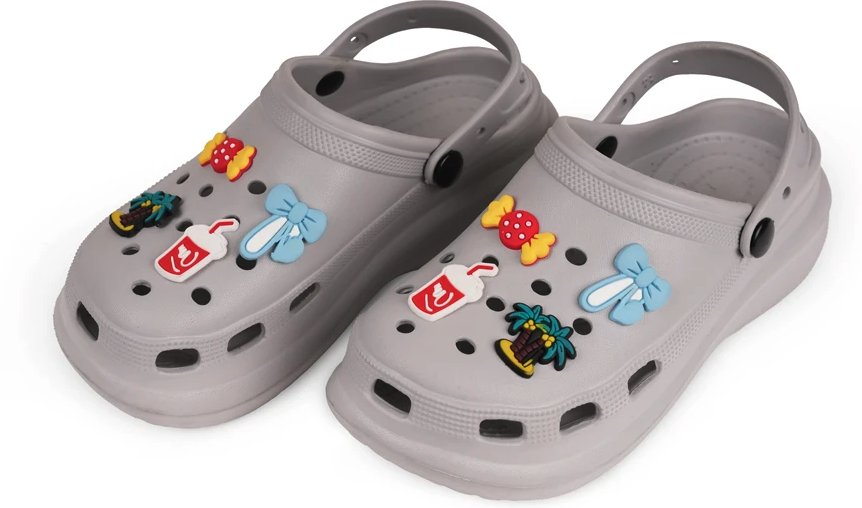} & \textbf{Attribute:} Pattern 

\textbf{Reason:} Based on the image, the shoe has a pattern on the heel, which is a floral print. Therefore, the most suitable value for heel\_pattern is 'Floral Print'. 

\textbf{Answer:} Floral Print & \textbf{Attribute:} Pattern 

\textbf{Reason:} Based on the image, the shoe has a solid pattern with colorful designs on the heel. The designs are not intricate or detailed, but rather simple and playful, which aligns with the 'Solid' pattern option. 

\textbf{Answer:} Solid \\
\hline

\includegraphics[width=100pt, height = 120pt]{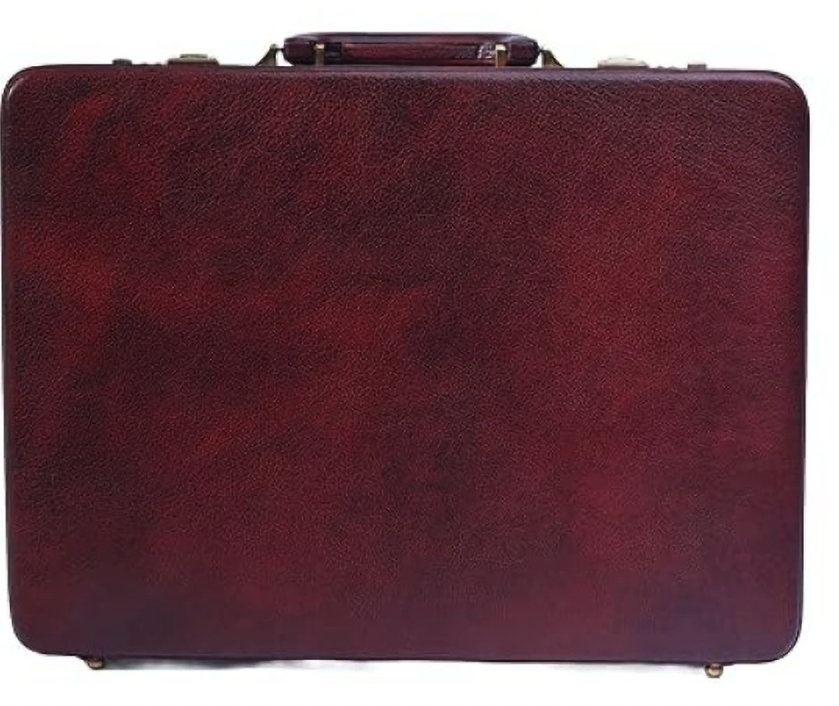}  \includegraphics[width=100pt, height = 120pt]{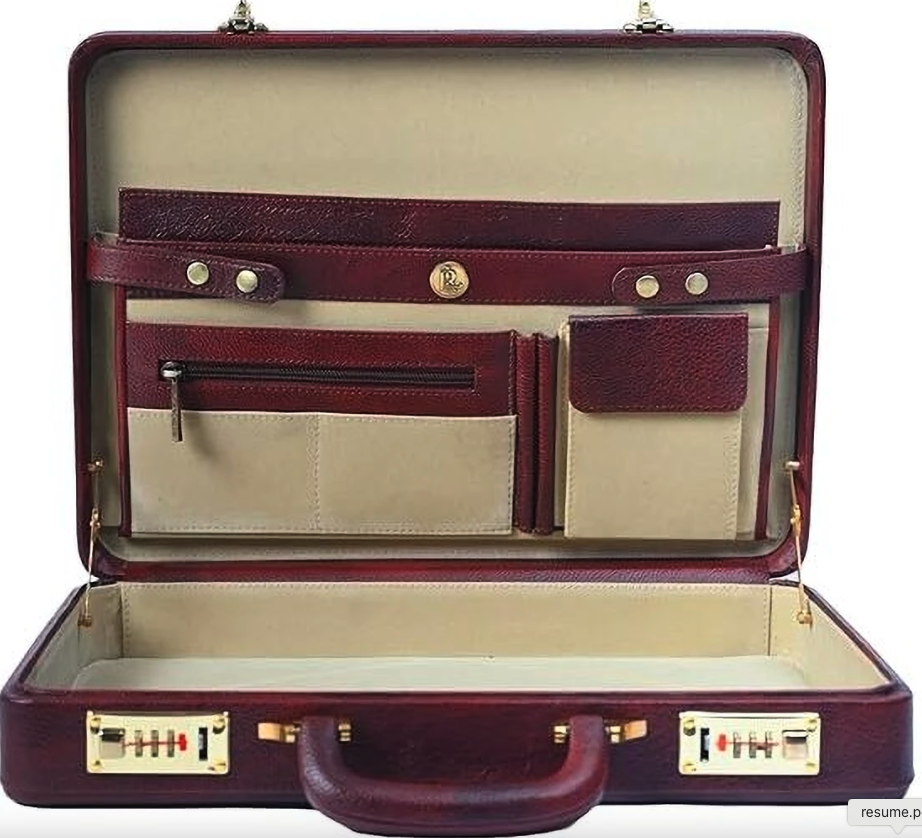} & \textbf{Attribute:} number\_of\_wheels 

\textbf{Reason:} Based on the image, the suitcase has two wheels. Therefore, the most suitable value for number\_of\_wheels is 2. 

\textbf{Answer:} 2 & \textbf{Attribute:} number\_of\_wheels 

\textbf{Reason:} Based on the image, the suitcase appears to be a traditional piece of luggage without any visible wheels. Most briefcases or suitcases do not have wheels. 

\textbf{Answer:} 0 \\

\hline

\end{tabular}
\caption{Sample results where the DPO training with unlabeled data improves the prediction accuracy of a Peft-10 SFT baseline.}
\label{tab:slbdpo}
\end{table*}

\section{Related works}

DPO \cite{rafailov2023direct} has been shown to align LLM outputs with human preferences in a stable and sample-efficient manner. 
Huang et al. \cite{huang2022largelanguagemodelsselfimprove} demonstrated the effectiveness of self-training where a high-capacity teacher model can generate high‐confidence pseudo‐labels on unlabeled corpora and iteratively refine a smaller student model, yielding substantial gains without additional human annotation.
Wang et al. \cite{wang2024selftrainingdirectpreferenceoptimization} demonstrated that self-training combined with DPO can substantially improve model accuracy on mathematical problem-solving tasks. In their method, multiple reasoning chains are generated for each problem, and external computation tools and rule-based logic are employed to identify  “winning” versus “losing” completions for DPO training. In contrast, our approach leverages self-consistency as an intrinsic correctness proxy to construct pairwise datasets for DPO training automatically. 
Consensus as labels has been used in the context of VLMs. Zhong et al. \cite{zhong2024vlmcplconsensuspseudolabels} introduced a human–annotation–free approach that leverages zero-shot inference from pre-trained VLMs. Their VLM-CPL framework generates both prompt-based and feature-based pseudo labels, applies a prompt–feature consensus filter to select reliable samples, and employs high-confidence cross supervision to discard noisy labels and perform semi-supervised learning.

\section{Conclusion}

In this paper, we demonstrated that leveraging unlabeled product listings through DPO can yield substantial accuracy gains for small VLMs on the e-commerce attribute prediction task. By employing self-consistency to construct preferred/dispreferred pairs from zero-shot model generations automatically, we eliminate the need for costly manual annotations or external validators. Furthermore, we utilize PEFT fine-tuning, which confines updates to low-rank adapter weights, thereby reducing computation and memory overhead. We also explored a self-learning paradigm driven by pseudo-labels; however, this approach consistently degraded performance, highlighting the risk of confirmation bias when small models iteratively fine-tune their noisy outputs. We performed an in-depth analysis of how the self-consistency threshold and the volume of unlabeled data affect both DPO and self-learning. We observed that, unlike self-learning, DPO training continues to improve as the amount of unlabeled data increases, confirming that large unlabeled corpora can be effectively harnessed to enhance small VLMs without sacrificing stability. The proposed unlabeled DPO training can be used to improve any SFT VLM model with objective  outputs.

{
    \small
    \bibliographystyle{ieeenat_fullname}
    \bibliography{main}
}


\clearpage




\end{document}